\pdfoutput=1

\documentclass[11pt]{article}

\usepackage[final]{acl}

\usepackage{times}
\usepackage{latexsym}

\usepackage[T1]{fontenc}

\usepackage[utf8]{inputenc}

\usepackage{microtype}

\usepackage{inconsolata}

\usepackage{graphicx}
\usepackage{booktabs}
\usepackage{CJKutf8}
\usepackage{multirow}
\usepackage{amsmath, amsfonts}
\usepackage{hyperref}
\usepackage{bbm}
\usepackage{afterpage}
\usepackage{placeins}
\usepackage{float}

\title{Do Large Language Models Have an English ``Accent''? \\ Evaluating and Improving the Naturalness of Multilingual LLMs}

 \author{Yanzhu Guo$^{2,3}$\Thanks{Work done during internship at Apple.},  Simone Conia$^{4}$, Zelin Zhou$^{1}$, Min Li$^{1}$, Saloni Potdar$^{1}$, Henry Xiao$^{1}$\\
 $^1$Apple
 $^2$Inria Paris
 $^3$École Polytechnique 
 $^4$Sapienza University of Rome
 \\ }

\begin{document}

\pagestyle{plain}
\thispagestyle{plain}

\maketitle
\begin{abstract}
Current Large Language Models (LLMs) are predominantly designed with English as the primary language, and even the few that are multilingual tend to exhibit strong English-centric biases. Much like speakers who might produce awkward expressions when learning a second language, LLMs often generate unnatural outputs in non-English languages, reflecting English-centric patterns in both vocabulary and grammar. Despite the importance of this issue, the naturalness of multilingual LLM outputs has received limited attention. In this paper, we address this gap by introducing novel automatic corpus-level metrics to assess the lexical and syntactic naturalness of LLM outputs in a multilingual context. Using our new metrics, we evaluate state-of-the-art LLMs on a curated benchmark in French and Chinese\footnote{Chinese in this paper refers to Mandarin written in Simplified Chinese characters.}, revealing a tendency towards English-influenced patterns. To mitigate this issue, we also propose a simple and effective alignment method to improve the naturalness of an LLM in a target language and domain, achieving consistent improvements in naturalness without compromising the performance on general-purpose benchmarks. Our work highlights the importance of developing multilingual metrics, resources and methods for the new wave of multilingual LLMs.
\end{abstract}

\section{Introduction}

LLMs are becoming an integral part of society with a visible impact on the general public~\citep{heikkila2023ai, ZieglerDonkers+2024+263+272, peterson2024ai}. 
However, there exists a significant disparity in how different languages are represented in LLMs, as popular models are primarily designed with English in mind.
While this is a well-known issue and more multilingual LLMs are being released, most of them are still English-dominated~\citep{wendler-etal-2024-llamas}. A prominent example is the Llama 3.1 series of models~\citep{dubey2024llama}, which are claimed to be state-of-the-art multilingual LLMs: these models are trained on 15T tokens, yet only 8\% of the training data is declared to be non-English. 

We could make an analogy between these multilingual LLMs and native English speakers who are trying to acquire a new language~\citep{groves-etal-2018-treat, devore2023assessing}. Their language notions are built in an English-centric system, and they inevitably bring traces of English habits into other languages when transferring their notions~\citep{papadimitriou-etal-2023-multilingual, wendler-etal-2024-llamas}.  
Moreover, because of the lack of data in non-English languages, multilingual LLMs are often exposed -- either during pre-training or post-training~\citep{yang2024qwen2, abdin2024phi} -- to texts translated from English. Both human and machine translated language are known to suffer from translationese artifacts, which set them apart from native content~\citep{bizzoni-etal-2020-human, luo-etal-2024-diverge}. LLMs trained on such data are susceptible to suffer from the same translationese problems.

In fact, even LLM-generated texts in English are known to exhibit distributional differences from human-written texts~\citep{guo-etal-2023-hc3, liang2024monitoring, liang2024mapping, shumailov2024ai}. Given the predominance of English data, this effect is likely more pronounced in non-English outputs. However, current evaluations of multilingual LLMs still focus on their task-solving capabilities~\citep{zheng2023judging, feng-etal-2023-factkb, zhang-etal-2024-safetybench, hendrycks2021measuring}, overlooking the aspect of naturalness. As LLMs increasingly influence various aspects of our lives, their tendency to produce less natural outputs in lower-resource languages in favor of English expressions could amplify the unfairness for the communities that speak these languages. 
Therefore, it is crucial to evaluate and improve the naturalness of multilingual LLMs to foster fair language representation.

In this paper, we take two steps towards this goal. 
Our first step is the introduction of a novel set of metrics to evaluate the naturalness of LLM outputs at a corpus level by comparing the lexical and syntactic distributions of LLM outputs with human written texts. 
We use these metrics, together with our new topically-aligned cross-lingual dataset, to benchmark and analyze the naturalness of state-of-the-art multilingual LLMs in English, French and Chinese. 
Our second step is the introduction of a simple and effective approach to enhancing the naturalness of multilingual LLMs in a target language. Using Direct Preference Optimization (DPO)~\citep{rafailov2023direct}, we leverage a new preference dataset that contrasts human-written responses with synthetically-manipulated ones. Experimental results show that our method consistently improves naturalness of an LLM in Chinese without sacrificing its capabilities on general-purpose benchmarks.

In summary, our contributions are threefold:
\begin{enumerate}
    \item We develop new metrics to evaluate the lexical and syntactic naturalness of LLM outputs in a multilingual setting;
    \item We create a benchmark for cross-lingual evaluation of LLM naturalness and draw insights from the benchmark results on important factors that could impact LLM naturalness;
    \item We propose an alignment approach for improving the naturalness of existing LLMs with promising results across models and domains.
\end{enumerate}
We hope our investigation will encourage further research on the limitations of multilingual LLMs beyond their scores on task-solving benchmarks.

\section{Related Work}

Although the evaluation of LLMs has received significant research interest in recent years, the greater part of this body of work has focused on aspects such as helpfulness~\citep{zheng2023judging}, factual accuracy~\citep{feng-etal-2023-factkb}, safety~\citep{zhang-etal-2024-safetybench}, fairness~\citep{chalkidis-etal-2022-fairlex}, and task-specific performance~\citep{hendrycks2021measuring}, leaving the naturalness of LLM outputs under-investigated. Naturalness is commonly used as an evaluation criterion in machine translation, but it has mostly relied on either human ratings~\citep{chen-etal-2024-iterative} or trained classifiers~\citep{liu-etal-2021-naturalness}. To the best of our knowledge, our work is the first to systematically investigate linguistic naturalness in multilingual LLM generations outside of the machine translation context. Nonetheless, several adjacent research areas are highly relevant to our focus, including translationese detection, linguistic diversity evaluation, and multilingual language model analysis.

\smallskip
\textbf{Translationese detection}, a well-established task in machine translation, aims to determine whether a text is originally written in the target language or translated from another language~\citep{volansky2015features, wintner-2016-translationese}. For instance, \citet{freitag-etal-2022-natural} employ a range of linguistic features -- including type-token ratio, lexical density, answer lengths, dependency tree height, constituency tree height, perplexity, etc. -- to train a classifier to distinguish between machine-translated and naturally occurring sentences. 
However, these classifiers are prone to overfit on specific training data.
In an effort to enhance the naturalness of machine translation outputs, \citet{freitag-etal-2022-natural} also tagged parallel training data based on target-side naturalness, contrasting models trained on natural versus translated text. Our approach builds on the concept of contrasting natural versus unnatural texts but avoids reliance on pre-trained classifiers. Instead, we use automatically manipulated unnatural texts and preference learning. We also extend the analysis beyond machine translation to general multilingual text generation.

\smallskip
\textbf{Linguistic diversity evaluation}~\citep{tevet-berant-2021-evaluating} is another area closely related to naturalness, as a key marker of unnaturalness in synthetic text is reduced linguistic diversity~\citep{vanmassenhove-etal-2021-machine, guo-etal-2024-curious}. Our work draws inspiration from the way these diversity features are computed. Past work compares the diversity of generated texts to human texts and consider texts to be natural if they approach the human level of diversity~\citep{freitag-etal-2022-natural}. However, such an approach primarily assesses the dispersion of the distribution while overlooking more holistic comparisons of linguistic features. To address this, we introduce metrics that directly compare vocabulary and syntactic distributions between human and machine-generated texts.

\smallskip
\textbf{Multilingual analysis} of LLMs has recently shown that models trained on unbalanced, English-heavy corpora often rely on English as an internal pivot language. For instance, \citet{wendler-etal-2024-llamas} demonstrate that the concept space of LLaMA-2 is more closely aligned with English than with other input languages. More empirically, \citet{papadimitriou-etal-2023-multilingual} show that multilingual BERT exhibits a bias toward English-like grammatical structures. Despite these findings, there has been no systematic study on how this English-centric tendency affects the linguistic naturalness of multilingual LLM outputs, particularly in open-ended downstream tasks.

\section{Evaluation Metrics for Naturalness}\label{sec:metrics}

In this section, we introduce a new set of evaluation metrics designed to assess the naturalness of multilingual text generation at the corpus level. While our approach also requires a reference set of natively written texts, it differs from widely used reference-based metrics such as BLEU~\citep{papineni-etal-2002-bleu}, ROUGE~\citep{lin-2004-rouge}, and BERTScore~\citep{bert-score}. These sample-level reference metrics often struggle to account for human label variability and uncertainty \citep{plank-2022-problem, giulianelli-etal-2023-comes}, particularly in open-ended tasks with multiple valid generations. Additionally, while a single text sample with certain choices of vocabulary or grammatical structure might seem natural, their repeated occurrence across many generations would raise a red flag \citep{guo-etal-2023-hc3}. Detecting unnaturalness in single instances can be difficult, but statistical patterns emerge more clearly at the corpus level, serving as stronger indicators \citep{liang2024monitoring, liang2024mapping}. Our metrics leverage these distributional patterns, offering a broader and more robust perspective on language use across large text collections.


Having highlighted the advantage of a distribution-level perspective, we propose a new definition for language model naturalness. Past studies have defined the naturalness of a single piece of text by asking \textit{``could it have been produced by a native speaker?''}~\citep{novikova-etal-2016-crowd, groves-etal-2018-treat, liu-etal-2021-naturalness}. We adapt this definition to the corpus level and define the naturalness of a language model by asking \textit{``could the set of texts generated by this language model have been produced by a group of native speakers?''}

Our metrics are inspired by divergence measures such as MAUVE \citep{pillutla2021mauve, pimentel2023on}, which quantify the information-theoretic divergence between the probability distributions of a language generator and a true natural language distribution. However, our approach differs by not relying on another language model to embed the generated texts. This reduces the risk of introducing intrinsic biases from the chosen embedding model, a crucial consideration in multilingual settings, where such the chosen embedding models are often English-dominated themselves. Our method is also more transparent and interpretable, as it clearly distinguish between two key aspects of linguistic naturalness: syntactic and lexical naturalness. However, our metrics focus exclusively on the linguistic form of the text and do not address semantic aspects. For analyzing semantics, the use of external embedding models may be inevitable.

In the following, we introduce the methodology for evaluating lexical and syntactic naturalness. The implementations of the metrics are described in Appendix \ref{sec:MetricsImplementation}.

\subsection{Lexical Naturalness}

We propose to measure the lexical naturalness of an LLM by comparing the vocabulary distribution of its generated text with that of human-written text.
More specifically, we put forward a lexical naturalness metric based on computing the \textbf{Jensen-Shannon Divergence (JSD)} between the lexical distributions of LLM-generated and human-written text from the same prompts.
The JSD provides a symmetric and bounded measure of difference between two distributions without them necessarily sharing the same support, i.e., the tokens in the vocabulary of the LLM in our case.\footnote{We process vocabularies at the word level instead of subword/token level to represent more meaningful lexical units.} Given the vocabulary distributions $P$ and $Q$ corresponding to human and model outputs, respectively, the JSD is calculated as follows:
$$
\text{JSD}(P || Q) = \frac{1}{2} \left( D_{\text{KL}}(P || M) + D_{\text{KL}}(Q || M) \right),
$$
where $M = \frac{1}{2}(P + Q)$ is the midpoint distribution, and $D_{\text{KL}}$ is the Kullback-Leibler divergence. By assessing the divergence, we can quantify how closely the model's vocabulary distribution aligns with human language, where lower values indicate greater similarity and higher lexical naturalness.
We note that our approach to directly compare the two vocabulary distributions implicitly captures the lexical statistical tendencies of past work that used type token-ratio and rank-frequency coefficient~\citep{meister-cotterell-2021-language}.

\subsection{Syntactic Naturalness}
To measure the syntactic naturalness of LLMs generations we leverage the \textbf{Universal Dependencies (UD)} grammar framework~\citep{nivre-etal-2020-universal}. UD provides well-defined, theoretically-grounded linguistic structures across different languages, making it suitable for cross-lingual comparisons.
Our proposed metric is based on representing each sentence as a dependency tree, where nodes correspond to words and edges specify the dependency relations between them. Additionally, each word is annotated with its corresponding Part-of-Speech (POS) tag as the node label.
At a high-level, our metric computes the structural similarity of all pairs of sentences in a corpus, clustering the ones who share similar syntax, allowing us to determine if there is a distributional difference between two groups (LLM-generated text and human-written text, in our case).

To compute the structural similarity between pairs of dependency trees, we propose to use the \textbf{Weisfeiler-Lehman (WL)} graph kernel \citep{shervashidze2011weisfeiler}. The WL kernel iteratively refines node labels based on the labels of neighboring nodes, thereby constructing a hierarchical encoding of the graph structure.
More formally, let $T_1$ and $T_2$ be two dependency trees with respective vertex, i.e., word, sets $V_1$ and $V_2$. The WL kernel, $K_{\text{WL}}(T_1, T_2)$, measures the similarity between $T_1$ and $T_2$ by comparing their subtree structures. Initially, each node in the trees is assigned a label based on its original POS tag. Then, at each iteration \( h \), we aggregate the labels of the node's neighbors into a multiset, which is then hashed into a new unique label. This process continues for $H$ iterations, and the kernel value is computed as the number of matching node labels across all iterations:
$$K_{\text{WL}}(T_1, T_2) = \sum_{h=0}^{H} \sum_{(v_1,v_2)\in(V_1,V_2)} \!\!\!\!\!\!\!\!\delta(\ell_h(v_1), \ell_h(v_2)),  
$$
where $\ell_h(v)$ is the label of node  $v$ at iteration $h$, and $\delta(\cdot, \cdot)$ is the Kronecker delta function. For our experiments, we fixed the number of iterations $H$ to 2, as this proved to be the most effective hyperparameter across various discriminative tasks for graphs \citep{shervashidze2011weisfeiler}.


Given a set of human-generated sentences $ \{s_i^{\text{h}}\}_{i=1}^{N_{\text{h}}}$ and model-generated sentences  $\{s_j^{\text{m}}\}_{j=1}^{N_{\text{m}}}$, we compute the WL kernel similarity between all pairs of dependency trees from these sets. The resulting kernel matrix $\mathbf{K} \in \mathbb{R}^{N_{\text{h}} \times N_{\text{m}}}$ has elements $K_{ij} = K_{\text{WL}}(T_i^{\text{h}}, T_j^{\text{m}})$, where $T_i^{\text{h}}$ and $T_j^{\text{m}}$ represent the dependency trees corresponding to sentences $s_i^{\text{h}}$ and $s_j^{\text{m}}$, respectively. 
This kernel matrix $\mathbf{K}$ captures structural similarities between the dependency trees of human-written and LLM-generated sentences.

\begin{table*}[ht]
\centering
\resizebox{\textwidth}{!}{%
\begin{tabular}{@{}llccccccc@{}}
\toprule
\multicolumn{2}{l}{} &
  \textbf{Human} &
  \textbf{Qwen1.5} &
  \textbf{Qwen2} &
  \textbf{Mistral-v0.3} &
  \textbf{Mistral-Nemo} &
  \textbf{Llama-3} &
  \textbf{Llama-3.1} \\ 
\multicolumn{2}{r}{\textbf{Model Size} (\# parameters)} &
  --- &
  7B &
  7B &
  7B &
  12B &
  8B &
  8B \\ \midrule
 &
  Lexical Divergence &
  \textbf{23.07} &
  {\color[HTML]{F56B00} \textbf{30.36}} &
  25.31 &
  {\color[HTML]{3531FF} \textbf{23.30}} &
  25.12 &
  29.00 &
  26.79 \\
\multirow{-2}{*}{\textbf{English}} &
  Syntactic Divergence &
  \textbf{~~3.53} &
  {\color[HTML]{F56B00} \textbf{22.19}} &
  13.67 &
  {\color[HTML]{3531FF} \textbf{13.56}} &
  14.77 &
  17.72 &
  16.80 \\ \midrule
 &
  Lexical Divergence &
  \textbf{25.91} &
  {\color[HTML]{F56B00} \textbf{41.00}} &
  37.08 &
  39.02 &
  34.78 &
  36.88 &
  {\color[HTML]{3531FF} \textbf{33.29}} \\
\multirow{-2}{*}{\textbf{Chinese}} &
  Syntactic Divergence &
  \textbf{~~2.93} &
  {\color[HTML]{F56B00} \textbf{23.33}} &
  20.66 &
  17.29 &
  12.84 &
  15.45 &
  {\color[HTML]{3531FF} \textbf{10.32}} \\ \midrule
 &
  Lexical Divergence &
  \textbf{24.25} &
  {\color[HTML]{F56B00} \textbf{38.35}} &
  31.18 &
  {\color[HTML]{3531FF} \textbf{28.73}} &
  31.34 &
  32.22 &
  31.52 \\
\multirow{-2}{*}{\textbf{French}} &
  Syntactic Divergence &
  \textbf{~~3.22} &
  {\color[HTML]{F56B00} \textbf{24.21}} &
  12.10 &
  12.72 &
  14.72 &
  17.88 &
  {\color[HTML]{3531FF} \textbf{11.27}} \\ \bottomrule
\end{tabular}%
}
\caption{Benchmark results for the lexical and syntactic naturalness of multilingual LLMs. All divergence values are presented as percentages and lower values indicate better naturalness. The best results for each language are highlighted in blue, while the worst are highlighted in orange.}
\label{tab:benchmark}
\end{table*}

Once the kernel matrix is obtained, we use the \textbf{Maximum Mean Discrepancy (MMD)} \citep{JMLR:v13:gretton12a} to compare the distributions of human-generated and model-generated sentences. In particular, the MMD$^2$ between the two sets of sentences is computed as:
$$
\frac{1}{N_{\text{h}}^2} \sum_{i, i'} K_{ii'} + \frac{1}{N_{\text{m}}^2} \sum_{j, j'} K_{jj'} - \frac{2}{N_{\text{h}} N_{\text{m}}} \sum_{i, j} K_{ij},
$$
where $K_{ii'}$ and $K_{jj'}$ represent the similarities within the human and model-generated sentence sets, and $K_{ij}$ represents the cross-similarities between the two sets. The resulting value provides a measure of syntactic divergence between the human-generated and model-generated sentences, with a lower value indicating greater similarity.

To conclude, our proposed syntactic naturalness metric quantifies the syntactic divergence between human and model-generated sentences by examining the distribution of their dependency tree structures. This approach considers both the dependency relationships and the hierarchical arrangement of substructures at multiple levels. Each step in our approach (e.g., the POS tagger, WL kernel, and MMD) has been externally validated for accuracy and effectiveness.


\section{Cross-lingual Analysis of Naturalness}

After introducing our new metrics, we proceed with a cross-lingual analysis of the naturalness of multilingual LLMs. This process involves curating a new dataset, selecting the models, and analyzing the benchmark results.

\subsection{Dataset and Evaluation}\label{subsec:dataset}

For our analysis, we require a corpus that is topically aligned across languages while preserving ground-truth naturalness. This means that the texts in each language must be natively written, not translated -- whether by humans or models. As a result, the parallel corpora typically used in machine translation research are not suitable. Instead, we construct a new dataset that satisfies our criteria starting from Wikipedia\footnote{Wikipedia is available under the CC BY-SA 4.0 license.}, which offers a wealth of texts that are frequently edited by native speakers and that we can topically-align across languages. 
Importantly, to minimize cultural bias in our dataset, we select the most-viewed Wikipedia entries across English, Chinese and French.\footnote{We use the Wikipedia API to retrieve the top-20 most-viewed entires in each language daily from June 2022 to June 2025. We then take the per-language union of the entries and retain their the cross-language intersection.}
Our data curation process results in 3,722 Wikipedia entries, each accompanied by descriptions in the three target languages. We discuss the preprocessing of this dataset in Appendix~\ref{sec:wiki}.

For each entry in the dataset, we instruct the models to complete a straightforward task: generating a description for the given entry in each of the three languages. We select this task to avoid overly constraining the content of the language model outputs, allowing them to generate in a more natural and organic manner. 
We stress that our focus is on the overall distribution of vocabulary and grammatical structures rather than on individual outputs. 
The prompts and generation settings used in this task are provided in Appendix~\ref{sec:prompts}.

\begin{CJK}{UTF8}{gbsn}

\begin{table*}[ht]
\centering
\resizebox{\textwidth}{!}{%
\begin{tabular}{lllcccl}
\toprule
\multirow{2}{*}{\textbf{POS Pattern}} &
  \multicolumn{2}{c}{\textbf{$n$-grams from Llama-3.1}} &
  \multicolumn{3}{c}{\textbf{Frequency}} &
  \multicolumn{1}{c}{\multirow{2}{*}{\textbf{Explanation}}} \\ \cmidrule{2-6}
 &
  English &
  Chinese &
  \begin{tabular}[c]{@{}l@{}}Native \\ English\end{tabular} &
  \begin{tabular}[c]{@{}l@{}}Native \\ Chinese\end{tabular} &
  \begin{tabular}[c]{@{}l@{}}Llama-3.1\\ Chinese\end{tabular} &
  \multicolumn{1}{c}{} \\ \midrule
(ADJ, CCONJ, ADJ) &
  \begin{tabular}[c]{@{}l@{}}(blue, and, white)\\ (incoming, and, outgoing)\\ (eastern, and, north)\end{tabular} &
  \begin{tabular}[c]{@{}l@{}}(真诚, 和, 直接)\\ (正常, 和, 合法)\\ (血腥, 和, 黑暗)\end{tabular} &
  37 &
  4 &
  22 &
  \begin{tabular}[c]{@{}l@{}}English requires conjunctions when listing \\ adjectives, while Chinese often omits them.\end{tabular} \\ \midrule
(PRON, AUX, VERB) &
  \begin{tabular}[c]{@{}l@{}}(They, were, married)\\ (he, was, named)\\ (it, was, built)\end{tabular} &
  \begin{tabular}[c]{@{}l@{}}(他, 被, 释放)\\ (她, 被, 称为)\\ (他, 被, 任命)\end{tabular} &
  175 &
  10 &
  56 &
  \begin{tabular}[c]{@{}l@{}}English uses passive constructions with \\ auxiliary verbs more frequently than Chinese.\end{tabular} \\ \midrule
(ADP, DET, NOUN, ADP) &
  \begin{tabular}[c]{@{}l@{}}(During, the, boom, of)\\ (at, the, end, of)\\ (At, the, height, of)\end{tabular} &
  \begin{tabular}[c]{@{}l@{}}(在, 多部, 电影, 中)\\ (在, 此, 职位, 上)\\ (在, 此次, 比赛, 中)\end{tabular} &
  300 &
  3 &
  22 &
  \begin{tabular}[c]{@{}l@{}}English depends heavily on prepositions and \\ determiners to structure sentences, while Chinese\\ tends to simplify these relationships.\end{tabular} \\ \bottomrule
\end{tabular}%
}
\caption{Extracted syntactic patterns that show the influence of English structures on how Llama-3.1 generates Chinese text. All frequencies are based on 40,000 $n$-grams of the same $n$. The Chinese and English outputs provided are neither semantically aligned nor translations.}
\label{tab:examples}
\end{table*}

\end{CJK}

\subsection{Multilingual LLMs}\label{subsec:models}

We experiment with three families of open-weight LLMs: Llama \citep{dubey2024llama}, Qwen \citep{yang2024qwen2}, and Mistral \citep{jiang2023mistral}. These models are selected for their state-of-the-art performance on diverse English and multilingual benchmarks. Additionally, they are developed by teams from regions where English, Chinese, and French are the official languages, respectively. While the exact composition of the training data for each model is not publicly available, we speculate that all are predominantly English-centric, though we expect Qwen to include more Chinese data and Mistral to include more French data than the others. We benchmark the two most recent versions from each model family to track changes in naturalness performance over time. For all models, we test their moderately scaled versions, using open-source implementations from the Transformers library \citep{wolf-etal-2020-transformers}. We focus on the instruction-tuned versions of the models, as these are more commonly used in real-world applications compared to the base versions.

\subsection{Results and Analysis}

Table~\ref{tab:benchmark} presents the naturalness performance of all models based on our proposed metrics (introduced in Section~\ref{sec:metrics}). As expected, the results show a consistent improvement in naturalness across newer versions of LLMs compared to their earlier counterparts (e.g., from Qwen-1.5 to Qwen-2). However, our metrics also reveal a persistent gap between the naturalness of human-written and LLM-generated texts, especially in non-English outputs, supporting our initial hypothesis.

Note that the human reference value is computed by randomly selecting non-overlapping subsets of human-written texts and measuring the divergence between them. These values are not obtained under the exact same conditions as the human-model divergences, which may involve overlapping prompts. This stricter setup makes the human reference value we report an upper bound on the human divergence baseline. As we show below, even this upper bound remains substantially lower than the human-model divergences.

\begin{figure}[ht]
    \centering
    \includegraphics[scale=0.38]{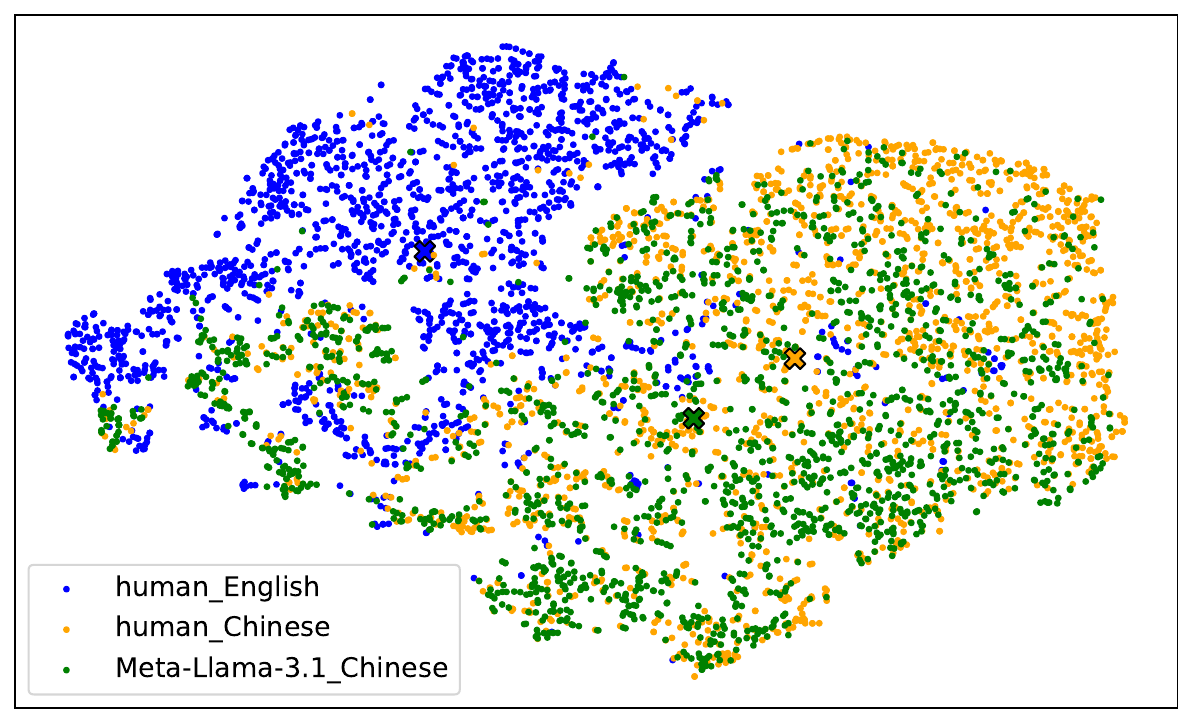}
    \caption{T-SNE visualization of the syntactic structures generated by Llama-3.1 in Chinese, compared to human-written Chinese and human-written English.}
    \label{fig:zh-en}
\end{figure}

\begin{figure*}[ht]
    \centering
    \includegraphics[width=0.8\textwidth]{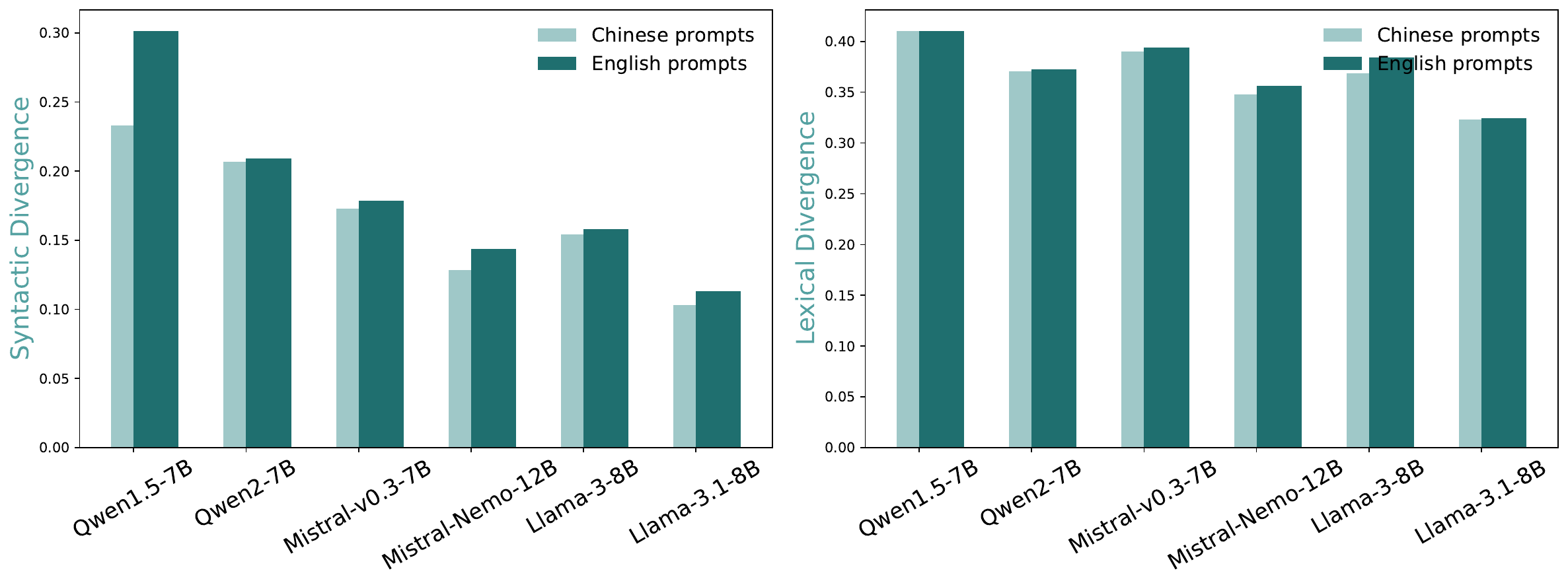}
    \caption{Comparison of divergence metrics for model generations in Chinese when the prompting language is either Chinese or English.}
    \label{fig:prompt_zh}
\end{figure*}

\begin{figure*}[ht]
    \centering
    \includegraphics[width=0.8\textwidth]{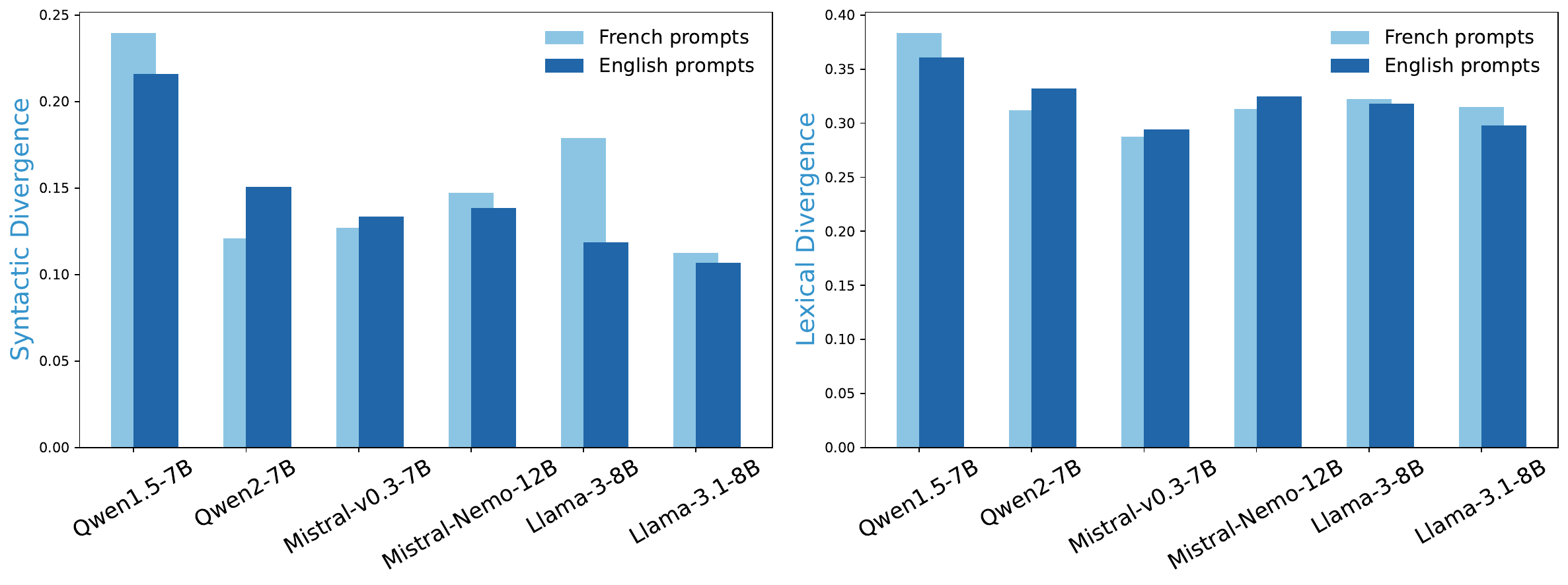}
    \caption{Comparison of divergence metrics for model generations in French when the prompting language is either French or English.}
    \label{fig:prompt_fr}
\end{figure*}

\begin{figure*}[h]
    \centering
    \includegraphics[width=0.9\textwidth]{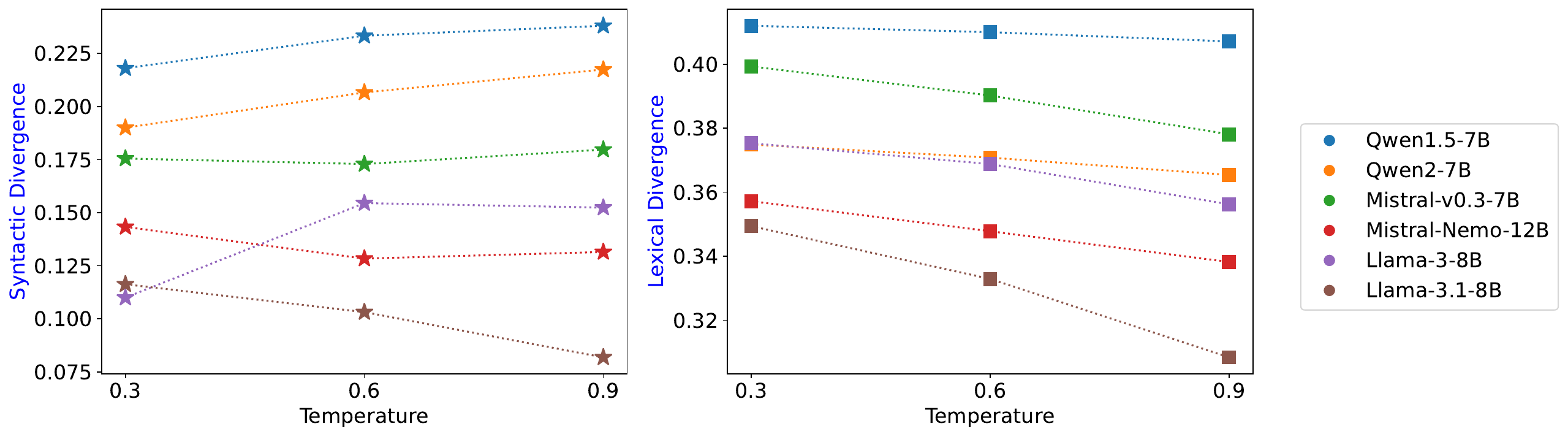}
    \caption{Variation of divergence metrics for model generations in Chinese when the decoding temperature changes.}
    \label{fig:decoding}
\end{figure*}

\paragraph{Gap in lexical naturalness.}
In terms of lexical divergence, English outputs -- especially from Mistral-v0.3 -- approach human reference values, while larger gaps persist in Chinese and French, with Chinese showing the most significant difference. This indicates that the models are lexically less natural in languages other than English, and the naturalness gap is more pronounced for languages that are typologically more distant from English. The syntactic divergence values across different languages are not directly comparable. This is due to the fact that the UD grammar parses languages at varying levels of granularity, as evidenced by the lower human divergence value in Chinese compared to English or French.

\paragraph{English ``accent'' in syntactic structures.}
When examining syntactic divergence, all models in all languages show significant differences from the human reference values. Since dependency trees, when parsed using the UD grammar, share a common space across languages, we can perform cross-lingual comparisons of syntactic structures. In Figure \ref{fig:zh-en}, we show Llama-3.1's English ``accent'' in syntax when generating Chinese outputs. Although Llama-3.1 is overall closer to the human distribution than other LLMs (as shown in Table~\ref{tab:benchmark}), its syntactic structures still exhibit greater similarity to human-written English than to native Chinese, suggesting that English syntactic patterns influence its Chinese generations. We can observe a similar effect in French, even if less noticeable since English and French belong to linguistic families (West Germanic and Romance, respectively) that are closer. The English accent of Llama-3.1 can also be seen in our case study of unnatural language patterns in text generated by Llama-3.1 in Chinese. Table~\ref{tab:examples} presents examples of POS tag n-grams that frequently appear in both English and Llama-3.1-generated Chinese, but much less frequently in native Chinese.


\paragraph{Impact of pretraining data.}
Among the evaluated models, the Qwen series consistently underperforms in naturalness across all languages, including Chinese, despite expectations that it would perform better due to its development in China and likely greater access to Chinese pretraining data. In contrast, Llama consistently produces more natural outputs. We hypothesize that Qwen's reliance on synthetic data during pretraining~\citep{yang2024qwen2}, in contrast to Llama’s explicit avoidance of synthetic data~\citep{dubey2024llama}, may explain this discrepancy. Notably, Mistral-v0.3 achieves the highest naturalness in English, possibly due to being the least multilingual among the models evaluated. Additionally, Mistral models show strong naturalness in French, likely benefiting from the use of more authentic French data, given that the team behind it is based in France. Unfortunately, we lack detailed information about the exact composition of the training data for any of these models. We hope to see future releases of multilingual LLMs following the principles of OLMo \citep{groeneveld2024olmo}, where not only the model weights but also the training data and pipeline are made open source.

\noindent \textbf{Impact of prompting language.}
For the previous benchmark results, we prompt the models in the same language in which we expect them to generate. To evaluate whether the input language affects the naturalness of the generations, we conduct an ablation study in a cross-lingual setting. Specifically, we manually translate the Chinese and French prompts into English and feed these English prompts to the LLM, while still instructing it to generate responses in Chinese or French. In other words, only the instructions are modified, the target language of the output remains unchanged. The results for the Chinese and French generations are presented in Figures \ref{fig:prompt_zh} and \ref{fig:prompt_fr}. For Chinese outputs, prompting in Chinese consistently leads to better naturalness. However, for French outputs, the results vary across models. We hypothesize that this is influenced by the proportion of instruction-response pairs the models encountered in different languages during instruction tuning. Additionally, the closer typological relationship between French and English, compared to Chinese, may also contribute.

\begin{table*}[ht]
\centering
\resizebox{0.98\textwidth}{!}{%
\begin{tabular}{ll|ccc|ccc}
\toprule
 &                      & \multicolumn{3}{c|}{\textbf{Llama-3.1-8B}}     & \multicolumn{3}{c}{\textbf{Qwen2-7B}}            \\
 &                      & Unaligned & Llama data     & Qwen data      & Unaligned    & Llama data     & Qwen data      \\ \midrule
\multirow{3}{*}{\textbf{\begin{tabular}[c]{@{}l@{}}Essay \\ Generation\end{tabular}}} &
  Lexical Divergence \(\downarrow\)&
  34.89 &
  35.06 &
  \textbf{34.56} &
  41.45 &
  \textbf{40.58} &
  40.60 \\
 & Syntactic Divergence \(\downarrow\)& 12.63       & 12.59          & \textbf{11.75} & 22.66          & \textbf{21.46} & 22.41          \\
 & CMMLU Accuracy  \(\uparrow\)   & 55.38       & \textbf{55.62} & 55.61          & 80.08          & 80.29          & \textbf{80.36} \\ \midrule
\multirow{3}{*}{\textbf{\begin{tabular}[c]{@{}l@{}}Question \\ Answering\end{tabular}}} &
  Lexical Divergence \(\downarrow\) &
  28.05 &
  \textbf{28.01} &
  28.04 &
  28.09 &
  27.58 &
  \textbf{26.92} \\
 & Syntactic Divergence \(\downarrow\) & 12.19       & ~~\textbf{9.44}  & ~~9.79           & 12.25          & \textbf{10.78} & 11.19          \\
 & CMMLU Accuracy  \(\uparrow\)     & 55.38       & \textbf{55.47} & \textbf{55.47} & 80.08 & 80.47          & \textbf{80.49} \\ \bottomrule
\end{tabular}%
}
\caption{Naturalness alignment results for essay generation and question answering tasks in Chinese. Lower divergence values indicate better naturalness, while higher CMMLU accuracy reflects stronger performance.}
\label{tab:dpo_results}
\end{table*}

\begin{CJK}{UTF8}{gbsn}

\begin{table*}[ht]
\centering
\resizebox{\textwidth}{!}{%
\begin{tabular}{lll}
\hline
Theme of Prompt &
  Output &
   \\ \hline
\multirow{2}{*}{\begin{tabular}[c]{@{}l@{}}夹乒乓大赛\\ \\ \textit{The Ping-Pong} \\  \textit{Clamping Contest}\end{tabular}} &
  Unaligned &
  \begin{tabular}[c]{@{}l@{}}今天，老师组织了一场特殊的乒乓球比赛——夹乒乓大赛。<......> 最后，比赛结束了，\colorbox{orange}{胜利者被宣布出来}。\\ \textit{Today, the teacher organized a special ping-pong competition—the Ping-Pong Clamping Contest. <......> In the end, the} \\ \textit{competition concluded, and \textcolor{orange}{the winner was announced}}.\end{tabular} \\ \cline{2-3} 
 &
  Aligned &
  \begin{tabular}[c]{@{}l@{}}今天，老师突然宣布要举办一场夹乒乓大赛。<......>  最后，\colorbox{cyan}{夹得最快的人赢得了}夹乒乓大赛的冠军。\\ \textit{Today, the teacher suddenly announced that there would be a Ping-Pong Clamping Contest. <......> In the end, \textcolor{cyan}{the person}} \\ \textit{\textcolor{cyan}{who clamped the fastest won} the championship of the Ping-Pong Clamping Contest.}\end{tabular} \\ \hline
\end{tabular}%
}
\caption{Example of improved output from Llama-3.1 after applying naturalness alignment for Chinese essay generation. English translations are presented in italics. Unnatural expressions are marked in orange, while natural alternatives are highlighted in blue. Although the expression ``the winner was announced'' is common in English, its literal translation in Chinese is rarely used, despite being grammatically correct. It uses a passive construction which is far less common in Chinese and sounds overly formal. 
}
\label{tab:dpo_examples}
\end{table*}

\end{CJK}

\paragraph{Impact of decoding temperature.}
Previous benchmark results are generated using temperature sampling with $t=0.6$, which is found to provide the optimal balance between creativity and consistency for multilingual generation tasks \citep{dubey2024llama}. As an ablation study, we vary the decoding temperature between 0.3 and 0.9 to examine its impact on the naturalness of model outputs. Our results in Figure \ref{fig:decoding} demonstrate that for syntactic naturalness, models that are already natural become even better as the decoding temperature increases, whereas models with unnatural outputs show a further increase in unnaturalness. For vocabulary, all models show improved naturalness as the decoding temperature rises. It is also worth noting that the ranking of the models generally remains unchanged across different decoding settings.

\section{Improving Naturalness through Preference Tuning}

We now propose an approach to improve the naturalness of language model outputs. Among the various stages of LLM development, 
preference-based learning is the most effective for refining stylistic features~\cite{ivison2024unpacking}. Based on this, we focus on aligning models for better naturalness during the preference tuning stage. Among the available preference optimization methods, we opt for DPO due to its simplicity and efficiency.

\subsection{Preference Dataset Construction}

We construct the preference datasets starting with SFT (Supervised Fine-Tuning) datasets originally written in the target language. However, we find that non-translated, open-source SFT datasets for non-English languages are almost non-existent. For French, we are unable to locate any such datasets, and for Chinese -- despite being the second highest-resource language after English -- only a limited number is available. Consequently, we conduct our DPO experiments on Chinese, using two datasets that focus on essay generation and open-domain QA tasks. These tasks are well-suited for stylistic alignment due to their open-ended, creative nature. For essay generation, we use the ``composition'' split from the Firefly dataset \citep{Firefly}, and for open-domain QA, we use the OpenLabel-Chinese Conversations Dataset \footnote{Released under the Apache 2.0 license.} \citep{OL-CC}.

The original instructions from the SFT datasets are preserved, with the initial response serving as the preferred response. To generate the unnatural, i.e., rejected, response we apply synthetic manipulations through paraphrasing and back-translation via English. The prompts we use to generate the rejected responses are listed in Appendix~\ref{sec:prompts1}. We ensure semantic consistency by using BLEU~\citep{papineni-etal-2002-bleu} to filter out pairs of chosen and rejected responses with insufficient similarity. Moreover, to ensure that the linguistic style of the rejected responses differs significantly from the chosen ones, we also filter out pairs with overly high BLEU scores. Based on an empirical analysis of BLEU score distributions, we define the final threshold as $0.15 < \textsc{Bleu}(\text{Chosen}, \text{Rejected}) < 0.9$. We also filter out responses shorter than 10 words (often in the style of multiple-choice questions), as they do not convey enough lexical or syntactic style. The statistics of our two resulting preference datasets are presented in Table \ref{tab:statistics}.

\subsection{Experimental Setup}
We believe that it is beneficial to perform naturalness alignment on models that have already undergone SFT and previous rounds of preference tuning, since preference tuning is typically done iteratively~\citep{dubey2024llama}.
We conduct experiments with the instruction-tuned versions of Llama-3.1 and Qwen2, which demonstrated the highest and lowest performance, respectively, in naturalness for Chinese in Table~\ref{tab:benchmark}.
To increase GPU memory usage efficiency and to avoid catastrophic forgetting, we apply Low-Rank Adaptation (LoRA)~\citep{hu2022lora} in conjunction with DPO. Detailed hyperparameters for both LoRA and DPO are provided in Appendix~\ref{sec:hyperparameters}.

We explore two approaches: (1) self-alignment, where the preference-tuning dataset is generated by the same model being aligned (e.g., using Llama-3.1 to create training data for Llama-3.1), and (2) cross-model alignment, where one model generates the dataset used to align a different model (e.g., using Llama-3.1 to produce training data for Qwen2).

\subsection{Results}
The results in Table~\ref{tab:dpo_results} show consistent improvements in both lexical and syntactic naturalness across models and tasks after applying our alignment method. In Table~\ref{tab:dpo_examples}, we provide an example on the essay generation task, comparing responses generated by Llama-3.1 before and after naturalness alignment using the same prompts.

Although we experiment with generating rejected responses using both paraphrasing and back-translation, the latter consistently yields slightly better performance than the former. We believe this is due to back-translation being more effective at introducing translationese artifacts into the rejected responses. Therefore, the final results in Table~\ref{tab:dpo_results} are based on back-translation. 
However, there is no definitive conclusion on whether rejected responses should be generated using self-alignment or cross-model alignment, and future work could explore generating response pairs with a combination of different models.

In addition, we evaluate our aligned models on the Chinese Massive Multitask Language Understanding (CMMLU) benchmark~\citep{li2023cmmlu} to ensure that their general capabilities are not compromised after naturalness alignment. Our results show that our alignment method not only improves linguistic naturalness but also slightly enhances overall language understanding performance. Interestingly, we observe that naturalness does not always positively correlate with language understanding performance. For example, Qwen2, despite achieving much higher scores than Llama-3.1 on the CMMLU benchmark, demonstrates lower naturalness, potentially due to the heavy use of synthetic data. This highlights the need for future research to consider naturalness as a complementary metric alongside conventional benchmark scores.

\section{Conclusion}
In this paper, we address the naturalness challenges of LLMs, particularly in multilingual contexts, by conducting experiments in English, French and Chinese. We introduce two corpus-level metrics to quantify the naturalness of model output distributions: one focused on vocabulary (lexical naturalness) and the other on grammatical structure (syntactic naturalness). These metrics are interpretable and free from biases introduced by external embeddings. Using these metrics, we benchmark state-of-the-art multilingual LLMs and analyze how factors such as training data, prompting language, and decoding strategy influence the generated language. Our analysis provides valuable insights into the current multilingual LLM landscape, complementing traditional performance benchmarks for task-solving capabilities. Finally, we propose an alignment method using DPO and a synthetically manipulated preference dataset to enhance the naturalness of model outputs. Experiments show that our aligned models consistently improve in both lexical and syntactic naturalness.

\section*{Limitations}

Our experiments do not include any low-resource languages because our approach relies on the availability of ground-truth distributions from native, human-written data, which is often scarce or unavailable for many languages. While our naturalness evaluation metrics are designed to be language-agnostic, they depend on reliable word tokenizers and dependency parsers for the languages studied. This is discussed in greater detail in Appendix~\ref{app:adaptation}. Unfortunately, such tools are still lacking for most low-resource languages. However, we argue that naturalness evaluation should not be the priority for these languages at this stage, as it is only meaningful once models achieve a certain baseline level of overall performance.

Our cross-lingual benchmark was limited to the Wikipedia domain, as this domain provides topically aligned, natively written content across languages. Wikipedia was the only source we could access with the necessary data. Although Wikipedia text serves as a reasonable proxy for general domain text, it may not guarantee that our findings are applicable across other domains. In the future, it would be valuable to develop more non-translated cross-lingual corpora for additional domains and extend the naturalness evaluation to those areas as well.

Our alignment approach has so far only been tested on data from essay generation and general domain question-answering tasks. These tasks were chosen because their creative and open-ended nature allows for the expression of stylistic features. However, applying this method to more knowledge-intensive and constrained tasks may introduce unintended knowledge editing, potentially leading to increased hallucination risks. Furthermore, our experiments were limited to Chinese due to the lack of natively written SFT datasets in other non-English languages. The Aya dataset \citep{singh-etal-2024-aya}, while a valuable resource for multilingual instruction tuning with native data, provides too few samples in each language. For example, after filtering for response length, we obtained only 958 samples in French from Aya, which is insufficient for our alignment and evaluation approach.

Collecting human annotations for naturalness evaluation is challenging. We initially attempted to gather human annotations for a meta-evaluation of our metrics, but annotators reported difficulty in distinguishing linguistic naturalness from individual generations. Our naturalness evaluation operates on a corpus level. As discussed in Section \ref{sec:metrics}, while a single text with specific vocabulary choices or grammatical structures may appear natural, repeated occurrences of these features across many generations would raise concerns. Human evaluators cannot easily process a large corpus and identify these patterns. Therefore, we believe that our automatic metrics address this gap where human evaluations fall short.

Finally, our evaluation and alignment methods focus solely on the naturalness of linguistic form, without considering social biases. However, we believe linguistic biases are significantly underexplored compared to social biases, and we aim to bridge this gap by taking an initial step in this direction.

 \section*{Acknowledgments}

We would like to thank all the people at Apple who supported this work and provided helpful feedback. The majority of this work was carried out during Yanzhu Guo's internship at Apple. Simone Conia gratefully acknowledges the support of the PNRR MUR project PE0000013-FAIR, which fully funds his fellowship since October 2023.

\bibliography{anthology,custom}


\clearpage

\appendix

\begin{table*}[ht]
\centering
\resizebox{\textwidth}{!}{%
\renewcommand{\arraystretch}{1.5}
\begin{tabular}{lll}
\hline
\textbf{\begin{tabular}[c]{@{}l@{}}Target \\ Language\end{tabular}} &
  \textbf{\begin{tabular}[c]{@{}l@{}}Prompt \\ Language\end{tabular}} &
  \multicolumn{1}{c}{\textbf{Prompt}} \\ \hline
English                  & English & Please write a summary description of the follwing Wikipedia entry in English: <wiki entry in English>           \\ \hline
\multirow{2}{*}{Chinese} & Chinese & \begin{CJK}{UTF8}{gbsn}请用简体中文写出以下维基百科词条的摘要描述：\end{CJK}<wiki entry in Chinese>                                                                    \\
                         & English & Please write a summary description of the follwing Wikipedia page in Simplified Chinese: <wiki entry in Chinese> \\ \hline
\multirow{2}{*}{French} &
  French &
  Veuillez rédiger une description résumée de l'entrée Wikipedia suivante en français: <wiki entry in French> \\
                         & English & Please write a summary description of the following Wikipedia entry in French: <wiki entry in French>            \\ \hline
\end{tabular}%
}
\caption{Prompts for Wikipedia description generation. For generations in Chinese and French, we experiment with both prompts in the target language and in English, to study the impact of prompting language on naturalness.}
\label{tab:prompts}
\end{table*}

\begin{table*}[ht]
\centering
\resizebox{\textwidth}{!}{%
\begin{tabular}{@{}lll@{}}
\toprule
\multicolumn{2}{l}{\textbf{Transformation}} & \multicolumn{1}{c}{\textbf{Prompt}} \\ \midrule
\multirow{2}{*}{\begin{tabular}[c]{@{}l@{}}Back \\ Translation\end{tabular}} &
  zh->en &
  \begin{tabular}[c]{@{}l@{}}You are a professional translator. You always show the translated version, without any additional explanations or format changes. \\ \\ Translate from English into Simplified Chinese:\\ \\ <text in Chinese>\end{tabular} \\ \cmidrule(l){2-3} 
 &
  en->zh &
  \begin{tabular}[c]{@{}l@{}}You are a professional translator. You always show the translated version, without any additional explanations or format changes. \\ \\ Translate from Simplified Chinese into English:\\ \\ <text in English>\end{tabular} \\ \midrule
\multicolumn{2}{l}{Paraphrasing} &
  \begin{tabular}[c]{@{}l@{}}You are a professional editor who revises word choices and restructures sentences while preserving the original meaning. You \\ always show the edited version, without any additional explanations or format changes. \\ \\ Edit the following text:\\ \\ <text in Chinese>\end{tabular} \\ \bottomrule
\end{tabular}%
}
\caption{Prompts used to generate rejected responses in the preference tuning dataset.}
\label{tab:prompts_preference}
\end{table*}

\begin{table*}[ht]
\resizebox{\textwidth}{!}{%
\begin{tabular}{@{}llllllll@{}}
\toprule
                                  &         & Qwen1.5-7B & Qwen2-7B & Mistral-v0.3-7B & Mistral-Nemo-12B & Llama-3-8B & Llama-3.1-8B \\ \midrule
\multirow{2}{*}{\textbf{English}} & ROUGE-L & 14.98      & 17.81    & 15.36           & 18.49            & 17.52      & 18.35        \\
                                  & BLEU    & 1.01       & 2.68     & 1.21            & 2.94             & 2.91       & 4.93         \\
\multirow{2}{*}{\textbf{Chinese}} & ROUGE-L & 11.56      & 13.65    & 13.17           & 10.81            & 13.14      & 12.16        \\
                                  & BLEU    & 0.56       & 1.05     & 0.93            & 0.48             & 0.96       & 0.99         \\
\multirow{2}{*}{\textbf{French}}  & ROUGE-L & 13.28      & 16.88    & 16.56           & 16.14            & 16.78      & 15.88        \\
                                  & BLEU    & 0.67       & 2.40     & 2.69            & 1.91             & 3.58       & 3.25         \\ \bottomrule
\end{tabular}%
}
\caption{BLEU and ROUGE scores for the models evaluated in the naturalness benchmark.}
\label{tab:bleu-rouge}
\end{table*}

\begin{table*}[h]
    \centering
    \resizebox{0.8\textwidth}{!}{%
    \begin{tabular}{ccccclc}
        \toprule
         &  \#Train&  \#Test& Len$_{prompt}$ &Len$_{chosen}$ &Len$_{rejected}$& 
         BLEU($chosen, rejected$)\\
         \midrule
         Essay&  10,050&  4,950& 29  & 564 & 523 & 0.29 \\
         QA&  4,281 &  2,109& 52  & 185 & 176 & 0.46\\
        \bottomrule
    \end{tabular}
    }
    \caption{Statistics of our preference tuning datasets constructed with back translation. All tokenization for length calculation was performed using the Llama-3.1 tokenizer.}
    \label{tab:statistics}
\end{table*}

\begin{table*}[ht!]
\centering
\resizebox{\textwidth}{!}{%
\begin{tabular}{@{}llllllllll@{}}
\toprule
                   & \multicolumn{4}{l}{\textbf{General}}                                        & \multicolumn{4}{l}{\textbf{LORA}}                   & \textbf{DPO} \\
\textbf{Parameter} & learning\_rate & max\_grad\_norm & warmup\_ratio & per\_device\_batch\_size & lora\_alpha & lora\_dropout & r   & target\_modules & beta         \\ \midrule
\textbf{Value}     & 5e-6           & 0.3             & 0.1           & 6                        & 128         & 0.05          & 256 & all-linear      & 0.5          \\ \bottomrule
\end{tabular}%
}
\caption{Hyperparameters used for preference tuning. Initial values are based on recommendations from the DPO and LoRA papers, with minimal additional tuning applied.}
\label{tab:hyper}
\end{table*}

\section{Preprocessing of the Wikipedia Dataset}
\label{sec:wiki}
We truncate any summaries exceeding 1024 tokens, as determined by Llama-3.1's tokenizer. Since the Chinese version of Wikipedia often contains content in traditional Chinese, we use the zhconv-rs\footnote{\url{https://github.com/Gowee/zhconv-rs}} library to convert all text to simplified Chinese for consistency during pre-processing.

\section{Wikipedia Description Generation}
\label{sec:prompts}
We now present the employed prompts and settings for Wikipedia Description Generation.

\subsection{Prompts} 

The prompts used in our experiments are listed in Table \ref{tab:prompts}. These prompts are then formatted using the default chat template\footnote{\url{https://huggingface.co/docs/transformers/main/en/chat_templating}} provided with each model in the Transformers library. We intentionally keep the prompts simple and do not include in-context learning to ensure the model’s behavior remains generalizable. After manually reviewing 50 generations from each model, we confirm that all models can follow the instructions and produce outputs aligned with our expectations. While the generated texts may not always be completely factual, they are consistently fluent and relevant to the given prompt, making them suitable for analyzing the general linguistic patterns of language models. Additionally, we use the fasttext-langdetect library\footnote{\url{https://pypi.org/project/fasttext-langdetect/}} to identify the language of the generations, filtering out those that are not in the target language. Across all models and languages, more than 99\% of the generations are correctly classified in the target language.

\subsection{Generation Settings} We use bf16 precision for generation, with a max\_new\_tokens limit of 1024, matching the truncation length of the human-written summaries. For most experiments, except those analyzing the effect of decoding temperature, we use temperature sampling with a temperature of 0.6 and a repetition\_penalty of 1.02.

\section{Implementation of Naturalness Metrics}\label{sec:MetricsImplementation}

For lexical naturalness, we process vocabulary at the word level instead of subword token level to represent more meaningful lexical units. We use the Jieba\footnote{\url{https://github.com/fxsjy/jieba}} tokenizer for Chinese and the NLTK\footnote{\url{https://www.nltk.org}} tokenizer for French and English. We remove punctuation and digits, but do not discard stop words as they are an important linguistic feature \citep{meister-cotterell-2021-language}.

For syntactic divergence, we parse sentences into dependency trees using the Stanza toolkit\footnote{Released under the ODC-By v1.0 license.} \citep{qi-etal-2020-stanza}, which generates dependency trees for each sentence according to the UD grammar \citep{nivre-etal-2020-universal}. We use the implementation of the WL kernel in the GraKeL library \citep{JMLR:v21:18-370} and normalize all kernel values between 0 and 1.

The divergence values are calculated between large distributions (60K words for lexical divergence and 3K sentences for syntactic divergence), so the measures are relatively stable. We try bootstrapping with 10 different randomizations for each measure and find the variation interval to be within 5\% for both.

\section{Comparison with N-gram Overlap Metrics}

We compute BLEU and ROUGE scores for the models evaluated in the naturalness benchmark (Table~\ref{tab:benchmark}) and report them in Table~\ref{tab:bleu-rouge}. These n-gram overlap metrics show markedly different behavior from our naturalness metrics. For example, while our metrics consistently identify the Qwen models as the least natural across all three languages, this pattern is not reflected in their BLEU or ROUGE scores. Similarly, our metrics capture a steady improvement in naturalness in newer LLM versions compared to earlier ones, a trend that BLEU and ROUGE fail to detect.

\section{Preference Dataset Generation}

\label{sec:preference}

We provide the prompts used for generating our preference dataset, along with statistical insights into the resulting dataset.

\subsection{Prompts}

\label{sec:prompts1}

The prompts used for generation of the preference dataset are presented in Table \ref{tab:prompts_preference}. Here, we only use English prompts, as our previous experiments show that prompting in English degrades the naturalness of generated Chinese, helping us produce the desired unnatural responses. The generation setting is the same as for Wikipedia Description Generation.

\subsection{Dataset Statistics}

Statistics of the generated preference dataset is shown in Table \ref{tab:statistics}. Rejected responses are shorter than the chosen ones on average, which may be due to models' tendencies to prioritize words that occur more frequently in the tokenizer's training set, resulting in fewer subword splits.

\section{Hyperparameters for Preference Tuning}
\label{sec:hyperparameters}

We conduct experiments using 8 Nvidia A100 GPUs, each with 40GB of memory. All models are trained in bf16 precision for 1 epoch. DPO training was performed with data parallelism, taking approximately 2 to 3 hours per model per dataset. We utilize the DPO implementation from the trl library\footnote{\url{https://huggingface.co/docs/trl/index}} and the LORA implementation from the PEFT library\footnote{\url{https://huggingface.co/docs/peft/index}}. The best-performing hyperparameters are listed in Table \ref{tab:hyper}.

\section{Adaptation of Naturalness Metrics for Other Languages}
\label{app:adaptation}

In our current approach, the lexical naturalness metric operates at the word level, as words are meaningful lexical units in the three languages (English, French and Chinese) analyzed in this study. However, for polysynthetic and agglutinative languages, which feature richer morphological structures and sparser word distributions, morphemes may be more suitable as the basic lexical units. However, adapting our metrics for these languages would require morphological parsing, which might be challenging in some cases, especially for polysynthetic languages.
For the syntactic naturalness metric, we use POS units as defined by the UD grammar. In the languages we analyzed, these units typically align with individual words. However, in English and French, there exist multi-word tokens where a single orthographic token corresponds to multiple syntactic words, a detail which we will clarify in the final version of our paper. Extending our syntactic metrics to polysynthetic and agglutinative languages would similarly require using the specific annotation frameworks provided by the UD schema for each language. For instance, in Turkish, the UD schema also divides orthographic tokens into syntactic words.
We thank the reviewer for raising this interesting question. We believe that future efforts to extend our framework to additional languages should address not only the resource disparity between high-resource and low-resource languages but also integrate linguistically informed decisions to account for the diversity of typological features across languages.

\end{document}